\documentclass[10pt, a4paper]{article}

\usepackage[final]{lrec2026} 

\usepackage{tcolorbox}
\tcbuselibrary{skins, breakable}
\usepackage[table,xcdraw]{xcolor}
\usepackage{tabularray}
\UseTblrLibrary{booktabs}
\usepackage{multirow}
\usepackage{colortbl}
\usepackage{enumitem}
\usepackage{soul}
\usepackage{contour}
\usepackage[normalem]{ulem}
\usepackage{makecell} 

\contourlength{1.2pt}
\renewcommand{\ul}[1]{%
  \uline{\phantom{#1}}%
  \llap{\contour{white}{#1}}%
}

\newcommand{\wino}{\textsc{WinoMT}\textsubscript{eus}}
\newcommand{\flores}{\textsc{FLORES}+Gender}
\newcommand{\male}{\textcolor{azure3}{\textbf{\texttt{M}}}}
\newcommand{\female}{\textcolor{purple5}{\textbf{\texttt{F}}}}

\title{Gender Bias in MT for a Genderless Language:\\New Benchmarks for Basque}

\name{Amaia Murillo, Olatz Perez-de-Viñaspre, Naiara Perez} 

\address{HiTZ Center - Ixa, University of the       Basque Country UPV/EHU \\
         \texttt{\{name.surname\}@ehu.eus}\\}

\abstract{
Large language models (LLMs) and machine translation (MT) systems are increasingly used in our daily lives, but their outputs can reproduce gender bias present in the training data. Most resources for evaluating such biases are designed for English and reflect its sociocultural context, which limits their applicability to other languages. This work addresses this gap by introducing two new datasets to evaluate gender bias in translations involving Basque, a low-resource and genderless language. \wino{} adapts the WinoMT benchmark to examine how gender-neutral Basque occupations are translated into gendered languages such as Spanish and French. \flores, in turn, extends the FLORES+ benchmark to assess whether translation quality varies when translating from gendered languages (Spanish and English) into Basque depending on the gender of the referent. We evaluate several general-purpose LLMs and open and proprietary MT systems. The results reveal a systematic preference for masculine forms and, in some models, a slightly higher quality for masculine referents. Overall, these findings show that gender bias is still deeply rooted in these models, and highlight the need to develop evaluation methods that consider both linguistic features and cultural context.\\ \newline }

\begin{document}

\maketitleabstract

\section{Introduction}
\label{sec:introduction}

Language technologies are often trained on huge amounts of uncurated data extracted from the Internet and, as a result, they can inherit stereotypes, misrepresentations, derogatory language and other demeaning behaviours that disproportionately affect vulnerable and marginalized communities \citep{gallegos2024biasfairnesslargelanguage}. When deployed in real-world applications, these systems can reproduce and even amplify such biases, leading to harmful social consequences \citep{10.1145/3617694.3623257}. For that reason, evaluating these biases is a key step toward developing fairer models.

Among the different forms of bias, gender bias has been one of the most extensively studied \citep{devinney2022theoriesgendernlpbias, Cignarella_2025}. 
However, most evaluation datasets have been developed in English and reflect the sociocultural context of English-speaking communities \citep{saralegi-zulaika-2025-basqbbq}. Since bias is shaped by culture, these resources cannot be directly applied to other languages, where stereotypes and social dynamics differ \citep{jin2024kobbqkoreanbiasbenchmark, oppong-etal-2025-examining}. Furthermore, these datasets rely on English-specific linguistic features and on cues that may not be applicable to typologically different languages. In fact, this form of bias is often evaluated through explicit gender markers---such as pronouns \citep{rudinger2018genderbiascoreferenceresolution} or other marked forms \citep{stanovsky2019evaluatinggenderbiasmachine}---but this approach cannot be used in the same manner for languages that lack grammatical gender \citep{corral-saralegi-2022-gender}. As a consequence, many languages remain underrepresented in bias research, and Basque, in particular, faces a scarcity of resources for evaluating gender bias, as a low-resource and genderless language.

\begin{figure}[t]
    \includegraphics[width=\linewidth,trim={20pt 10pt 20pt 20pt},clip]{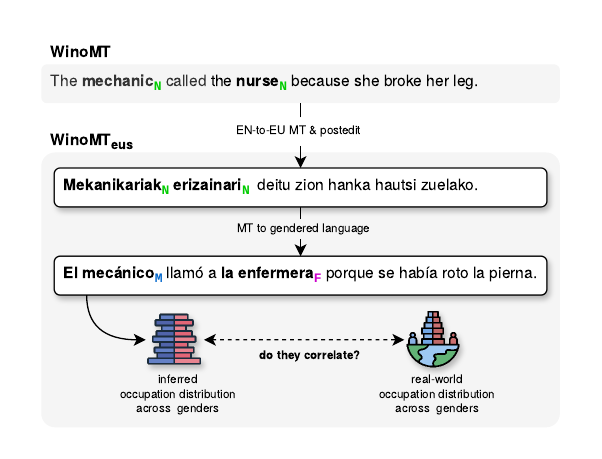}
    \caption{Overview of \wino{}. Sentences with gender-neutral (\textcolor{green7}{\textbf{\texttt{N}}}) occupation mentions, taken from WinoMT~\citep{stanovsky2019evaluatinggenderbiasmachine} and translated to Basque, are translated into a gendered language; we compare the resulting gendered occupation distribution (male \male{} or female \female{}) with real-world data.}
    \label{fig:winomteus}
\end{figure}

\begin{figure*}[t]
    \centering
    \includegraphics[width=\linewidth,trim={20pt 5pt 20pt 10pt},clip]{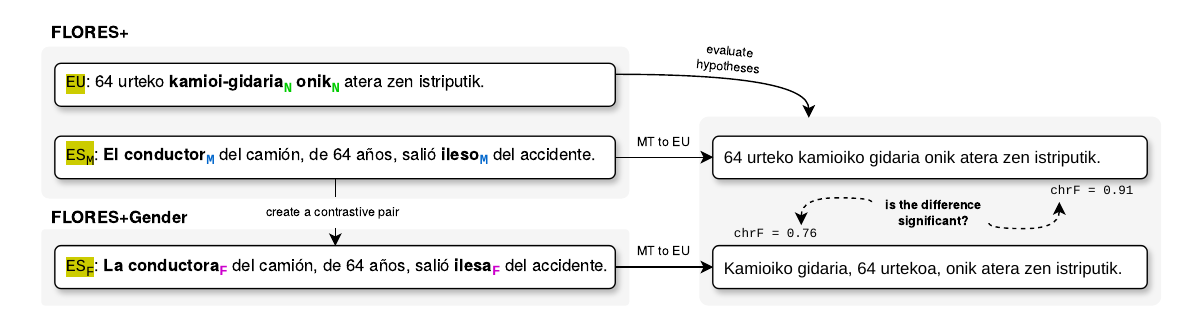}
    \caption{Overview of \flores. Using Basque-gendered language pairs from FLORES+,\protect\footnotemark[3] we create a contrastive version of the gendered sentence (male \male{} or female \female{}). Both are translated into Basque, and translation quality is compared to asses the effect of source-sentence gender.}
    \label{fig:floresgender}
\end{figure*}

To assess this type of bias in languages like Basque, machine translation (MT) offers a particularly useful framework. By translating between gendered and genderless languages, it is possible to observe how models handle gender information when explicit grammatical cues are present or absent \citep{zahraei-emami-2025-translate, rarrick2024gatexechallenge}. In short, translation not only reveals how gender disparities can manifest in practical scenarios, but also serves as a tool to uncover systematic gender preferences in the models.

In this work, we release two datasets designed to assess gender-related bias in Basque systems through MT. \textbf{\wino{}}\footnote{\url{https://github.com/amaiamurillo/winomteus}} (see Fig.~\ref{fig:winomteus}) focuses on how gender-neutral Basque occupations are rendered in gendered target languages, and, for the first time, whether they align with official employment statistics from the Basque Country. \textbf{\flores}\footnote{\url{https://github.com/amaiamurillo/flores_plus_gender}} (see Fig.~\ref{fig:floresgender}) studies whether translation quality varies when translating from gendered languages (Spanish and English) into Basque depending on the gender of the referent. Together, these resources provide a framework for evaluating gender bias in Basque MT and contribute to broadening bias evaluation beyond English.

\footnotetext[3]{\url{https://huggingface.co/datasets/openlanguagedata/flores_plus}}

\section{Background and Related Work}
\label{sec:background}

Gender bias in natural language processing (NLP) is a specific form of social bias that refers to unequal treatment or outcomes based on gender, often caused by historical and structural inequalities. It can manifest as intrinsic bias---encoded in the model's internal representations---or as extrinsic bias---which emerges in downstream tasks such as question answering or sentiment analysis~\citep{guo2024biaslargelanguagemodels}.
These inequities can result in representational and allocational harms, shaping how groups are depicted and how resources are distributed \citep{crawford2017trouble}. Evaluating these biases is crucial for building fairer language technologies.

\subsection{Bias Evaluation Resources}
\label{ssec:bias-evaluation-resources}

Most existing bias evaluation datasets to date have been created for English. These include Wino-style counterfactual benchmarks such as Winogender \citep{rudinger2018genderbiascoreferenceresolution} and WinoBias \citep{zhao2018genderbiascoreferenceresolution}, where pairs of minimally different sentences reveal stereotypical associations in model predictions, as well as other sentence-pair resources such as GAP \citep{webster-etal-2018-mind}, BUG \citep{levy-etal-2021-collecting-large} or BEC-Pro \citep{bartl-etal-2020-unmasking}. These resources have proved influential but remain tightly coupled to the grammatical and sociocultural properties of English, which features gendered pronouns and agreement patterns that do not generalize to typologically different languages. Prompt-based datasets (e.g., RealToxicityPrompts \citep{gehman2020realtoxicitypromptsevaluatingneuraltoxic}, BOLD \citep{Dhamala_2021}) have extended bias assessment to generative settings, yet such resources are again largely English-specific. Consequently, most low-resource and genderless languages lack dedicated benchmarks for gender bias evaluation.

\subsection{Basque and Existing Work}
\label{ssec:basque-and-existing-work}

Basque is a language isolate spoken by approximately three-quarters of a million people, most of whom reside in the Basque Country, a territory spanning areas of both Spain and France~\citep{soraluze-etal-2016-coreference}. Basque does not have grammatical gender, unlike its neighboring languages Spanish and French, and even personal pronouns are gender-neutral. Still, the absence of grammatical gender does not prevent the reproduction of bias, as language use itself can encode and transmit social stereotypes. Gender is conveyed through lexical choices: gender identity terms (\textit{mutil} `boy', \textit{emakume} `woman'), kinship terminology (\textit{izeba} `aunt', \textit{aitona} `grandfather'), occupations (\textit{andereño} `female schoolteacher', \textit{maisu} `male schoolteacher') and proper names (\textit{Aitor, Lander} and \textit{Asier} are typically associated with men, whereas \textit{Ane, Lide} and \textit{Nahia} are commonly given to women). 

Despite growing interest, resources for gender-bias evaluation in Basque remains scarce. The most closely related effort is the TANDO project~\citep{gete2022tando,gete2025tando}, which introduced contrastive Basque-Spanish test sets to examine gender selection and register phenomena in MT. In their setup, Basque source sentences were gender-neutral but required gendered translation to Spanish; translation quality was then compared across masculine and feminine references. TANDO therefore provided the first resource for assessing gender bias in Basque MT, specifically for the Basque-to-Spanish direction.

\subsection{Gender Bias Evaluation through MT}
\label{ssec:characteristics-of-basque}

MT offers a natural framework for evaluating gender bias in languages without grammatical gender. When translating from a genderless language (e.g., Basque or Turkish) into a gendered one (e.g., Spanish or English), models must make an explicit gender choice. Over many examples, a systematic preference for one gender can reveal underlying bias \citep{stanovsky2019evaluatinggenderbiasmachine, mastromichalakis2025assumedidentitiesquantifyinggender}. Conversely, translation from a gendered into genderless language allows researchers to measure whether translation quality differs by gendered input forms~\citep{costajussà2023multilingualholisticbiasextending}.

Building on this tradition, our work expands the scope of Basque MT bias evaluation in two ways. First, we present \textbf{\wino}, a counterfactual Wino-style dataset for translating \emph{from} Basque into gendered languages, where both masculine and feminine translations are plausible, and compare their distribution to official employment statistics, following \citet{mastromichalakis2025assumedidentitiesquantifyinggender}. Second, we introduce \textbf{\flores}, which extends FLORES+ to evaluate translations \emph{into} Basque from gendered languages (Spanish and English), testing whether gendered source forms affect translation quality. TANDO included Spanish to Basque but not English to Basque, making this the first multilingual setup of its kind for Basque.



\section{\wino{}}
\label{ssec:winomteus}

We translated and adapted WinoMT \cite{stanovsky2019evaluatinggenderbiasmachine} into Basque to examine how gender is assigned when translating gender-neutral occupations into gendered languages. The original dataset is a concatenation of Winogender \cite{rudinger2018genderbiascoreferenceresolution} and WinoBias \cite{zhao2018genderbiascoreferenceresolution}, which are two widely known template-based datasets for assessing occupational stereotypes through coreference resolution. Each instance includes two gender-neutral occupations and a pronoun that refers to one of them, and the model is required to resolve the ambiguity (see Fig.~\ref{fig:winomteus}). 

This task cannot be applied to Basque in the same way as in English, given that Basque lacks gendered pronouns. Hence, we translated the original dataset and created a Basque version, called \wino, to analyze how models render these professions when translating into gendered languages such as Spanish or French. The adaptation process involved five main stages:

\begin{enumerate}[noitemsep]
  \item \textbf{Glossary creation.} We created a glossary of 78 occupations to align English terms with culturally and linguistically equivalent forms in Basque. This process required dealing with issues such as the lack of lexical equivalents or professions without cultural counterparts.
  \item \textbf{Translation.} The original English dataset was translated into Basque using GPT-4o.
  \item \textbf{Post-editing and revision.} We post-edited the translated sentences while addressing the pitfalls identified by \citet{blodgett-etal-2021-stereotyping} present in the original dataset, including grammatical errors, sentence structure issues, logical failures and unnatural phrasing.
  \item \textbf{Cultural adaptation.} Contextually sensitive elements were adjusted to fit the Basque context (e.g., emergency numbers, currency and units of measurement).
  \item \textbf{Final filtering.} Duplicated sentences were removed, as several English sentences differ only in pronoun gender and therefore resulted in identical translations in Basque.
\end{enumerate}

\noindent The resulting dataset includes \textbf{1,827 sentences} in Basque with at least one occupation mention.

\section{\flores}
\label{ssec:flores-gender}

\flores{} is built on the FLORES+ benchmark \cite{nllb-24}, developed by Meta to assess MT systems for low-resource languages. The original dataset contains 1,012 sentences in English, evenly sampled from Wikinews, Wikijunior and Wikivoyage, translated into around 200 languages and varieties, including Basque, Spanish, Catalan, Valencian and Galician.

Adopting the approach of \citet{costajussà2023multilingualholisticbiasextending}, we reversed the translation direction to study whether the grammatical gender of the source text affects translation performance \emph{into} Basque. Specifically, we selected two source languages with different degrees of gender marking: Spanish, a strongly gendered language, and English, a weakly gendered one where gender is mostly conveyed through personal pronouns, possessive adjectives and gender-specific nouns.

For each source language, we construct contrastive versions of the dataset: one containing all the sentences in masculine form and another with the same sentences in feminine form. From the original \textbf{Spanish} set, we identified \textbf{363 sentences} with gendered references, and \textbf{155} in the \textbf{English} version. These were manually adapted to produce gender-controlled pairs while maintaining semantic equivalence. In addition, each sentence was manually annotated for
\begin{itemize}[noitemsep]
    \item \textbf{ME}: whether it contains mentions to multiple gendered human entities,
    \item \textbf{PN}: the presence of proper names, and
    \item \textbf{UM}: the presence of unmarked usage of the masculine form (applies only in Spanish).
\end{itemize}
\noindent These annotations, quantified in Tab.~\ref{tab:flores-gender}, support downstream analysis of whether such factors influence translation quality. 

\begin{table}[t]
    \centering
    \begin{tblr}{
        colspec={lrrr},
        cells={font=\footnotesize},
        row{1}={c,font=\footnotesize\bfseries},
        rowsep=1pt
    }
        \toprule
                   &  ME & PN & UM \\
        \midrule
         Spanish   & 24.52 & 19.65 & 60.01 \\
         English   & 29.03 & 49.68 & n/a \\
        \bottomrule
    \end{tblr}
    \caption{Quantification of \flores{} annotated phenomena (\% over total sentences).}
    \label{tab:flores-gender}
\end{table}

When adapting the data, we followed consistent criteria to ensure coherence and cultural plausibility. Proper names were replaced with alternatives of the opposite gender, prioritizing equivalents with similar spelling and cultural origin (e.g., \textit{Cristina}\textsubscript{ \female}/\textit{Cristian}\textsubscript{ \male}). In the case of highly recognizable names (e.g., \textit{Fernando Alonso}), we employed plausible alternatives whenever the context allows. For multi-entity sentences, we focused on modifying the main subject or the most contextually relevant gendered entity in the sentence instead. We also addressed various errors encountered during the annotation process. These included typos, unnecessary or missing words, as well as translation inconsistencies and gender agreement issues present in the original sets.

Finally, after producing the two gender-specific sets for each source language, we manually generated the two corresponding Basque sets with the same modifications, aimed at serving as a reference for automatic evaluation.

\section{Experimental Setup}
\label{sec:experimental-setup}

The goal of our experiments is to assess how gender bias manifests in MT involving Basque, using the two new resources introduced in this work. That is, we design complementary evaluations that examine gender bias both \textit{from} Basque into gendered target languages and \textit{into} Basque from other languages, using controlled datasets and quantitative metrics.

\subsection{Evaluation Methodology}
\label{ssec:evaluation-methodologyu}

This section describes the evaluation procedures employed to analyse gender bias in translation in both directions. We leverage the two datasets presented in \S\ref{ssec:winomteus} and \S\ref{ssec:flores-gender} to perform, respectively, \textit{i)} an analysis of how gender is expressed when translating \textit{from} Basque into gendered languages (Spanish and French), and \textit{ii)} an analysis of how translation \textit{into} Basque behaves with respect to gender-marked referents.

\subsubsection{\wino: \emph{from} Basque} 


This experiment analyses whether translation from Basque into a gendered language reinforces gender stereotypes or reflects actual labour distribution in the Basque Country. The evaluation consists in (see Fig.~\ref{fig:winomteus}) \textit{i)} machine-translating the dataset into Spanish and French, \textit{ii)} extracting the mentions of occupations in the translations and labelling their gender, and \textit{iii)} comparing the translated occupations' distribution to real-world labour statistics. In what follows, we detail the methodology employed to extract, gender-label, and evaluate the occupations obtained from the translated corpus.

\paragraph{Occupation extraction.} 
We automatically extracted the professions with Claude 4 Sonnet. For the prompt, we adopted a one-shot approach, giving the model an instruction to extract the occupations from the Spanish and French translations, along with one example showing how to perform the task (further details are available in Appendix A). In this process, we retrieved both the profession and the preceding article, and if no occupational term was detected, the output was marked as \texttt{[MISSING]}. To validate this extraction method, we manually checked 100 randomly selected sentences, and obtained 95.5\% of accuracy in occupation extraction. Next, we inferred the gender of each profession based on the gender of the article. In cases where the article was not enough to determine the gender, we relied on the morphological ending of the occupation (e.g., \textit{l'infirmi\ul{er}}\textsubscript{ \male}/\textit{l'infirmi\ul{ère}}\textsubscript{ \female} or \textit{l'administra\ul{teur}}\textsubscript{ \male}/\textit{l'administra\ul{trice}}\textsubscript{ \female}).


\paragraph{Labour statistics.} To evaluate the extent to which model biases reflect real-world disparities, we follow prior work by \citet{mastromichalakis2025assumedidentitiesquantifyinggender}, \citet{ibaraki-etal-2024-analyzing} and \citet{touileb-etal-2022-occupational}, comparing our results with actual labour distribution statistics. These statistics for the Basque Country were retrieved from Lanbide,\footnote{\url{https://www.lanbide.euskadi.eus}} the public employment service, as of July 2025. We manually mapped the profession categories provided by Lanbide to our glossary (see \S\ref{ssec:winomteus}) in order to identify the closest corresponding occupations. This allowed us to conduct two complementary evaluations: one at the aggregate level, comparing the overall distribution of masculine and feminine translations with labour statistics, and another at the profession level, analysing correlations for each occupation individually.

\paragraph{Evaluation Metrics.} On the one hand, we computed the \textbf{Pearson correlation coefficient} between the percentages of feminine and masculine translations for each occupation and the corresponding labour statistics to evaluate how closely the gender proportions in the model translations align with real-world data. On the other hand, we applied the \textbf{GRAPE} (\textit{Gender RAtion ProbabilitiEs}) metric, introduced by \citet{mastromichalakis2025assumedidentitiesquantifyinggender}, to carry out a more detailed analysis at profession level. This metric measures gender bias by comparing how frequently a model generates masculine or feminine forms relative to a reference distribution. It quantifies both the direction of the bias (GRAPE-M for masculine, GRAPE-F for feminine) and its magnitude (how strong that tendency is), starting from -1. For instance, a GRAPE-F value of 0.0 means that the system generates the same proportion of feminine forms as the reference distribution, whereas a value of 1.0 means that feminine forms are produced twice as often as expected. 

\subsubsection{\flores: \emph{into} Basque} 

Adopting the methodology of \citet{costajussà2023multilingualholisticbiasextending}, we reversed the translation direction to analyse whether the translation quality varies depending on the gender of the referent in the source sentence. For this purpose, we translated into Basque the English and Spanish versions of the \flores{} dataset.  We first evaluated overall translation quality across masculine and feminine subsets to detect global trends. Subsequently, we exploited the dataset’s fine-grained annotations to disaggregate results by the linguistic and contextual factors explained in \S\ref{ssec:flores-gender} and Tab.~\ref{tab:flores-gender}.

\paragraph{Evaluation metrics.} We compared the translations to their corresponding references employing standard automatic metrics using the SacreBLEU toolkit \cite{post2018call}: character-level F-score~\citep[chrF++;][]{popovic-2017-chrf} and Translation Edit Rate~\citep[TER;][]{snover-etal-2006-study}. To determine whether the observed differences between masculine and feminine subsets were statistically significant, we applied the paired approximate randomization test \citep{riezler-maxwell-2005-pitfalls} with 10,000 random permutations and significance levels $p<0.01$, $p<0.05$, and $p<0.1$.

\subsection{Models and Inference Setup}
\label{sec:models}

With the above described datasets, we assessed how gender bias manifests in MT across three different technical paradigms.

\paragraph{General-purpose LLMs.} In this category, we evaluated \textbf{Latxa 3.1 8B Instruct}\footnote{\url{https://huggingface.co/HiTZ/Latxa-Llama-3.1-8B-Instruct}} and \textbf{Latxa 3.1 70B Instruct}\footnote{\url{https://huggingface.co/HiTZ/Latxa-Llama-3.1-70B-Instruct}}, two general-purpose instructed LLMs for Basque developed by \citet{sainz2025instructing}. Both are based on \textbf{Llama 3.1 8B Instruct}\footnote{\url{https://huggingface.co/meta-llama/Llama-3.1-8B-Instruct}} and \textbf{Llama 3.1 70B Instruct}\footnote{\url{https://huggingface.co/meta-llama/Llama-3.1-70B-Instruct}}~\cite{grattafiori2024llama3herdmodels}, which were also included in the evaluation. All four models were deployed with vLLM~\citep{kwon2023efficient}, using a single A100 GPU for the 8B variants and two GPUs for 70B ones. We used the models' default generation settings with the temperature fixed at \texttt{0} to ensure reproducible results.

Additionally, we also assessed \textbf{GPT-5}\footnote{\url{https://cdn.openai.com/gpt-5-system-card.pdf}}, accessed via OpenAI API, with reasoning effort set to \texttt{minimal}. \textbf{Claude Sonnet 4}\footnote{\url{https://www.anthropic.com/news/claude-4}} was evaluated using Anthropic's API with the temperature also fixed at \texttt{0}. The last model in this category, \textbf{DeepSeek-V3.2-Exp Non-thinking}~\cite{deepseekai2025deepseekv3technicalreport}, was run via its API using default generation parameters. 

Beyond general-purpose models, we also included \textbf{SalamandraTA 7B Instruct}\footnote{\url{https://huggingface.co/BSC-LT/salamandra-7b-instruct}} \cite{gilabert2025salamandrasalamandratabscsubmission}, an instruction-tuned translation model derived from SalamandraTA 7B Base that supports 35 European languages, among them Basque, Spanish, English, French, Aragonese, Asturian and Catalan. The system and user prompts used across all models are provided in Appendix A. 

    

\paragraph{Open NMT models.} Among open neural machine translation (NMT) models, we tested  \textbf{MADLAD-400-3B-MT}\footnote{\url{https://huggingface.co/google/madlad400-3b-mt}} \cite{kudugunta2023madlad400}, a multilingual model covering 400 languages; \textbf{NLLB-200's 3.3B variant}\footnote{\url{https://huggingface.co/facebook/nllb-200-3.3B}} \cite{nllb-24}, designed to support low-resource languages and supports 200 language pairs; and \textbf{HiTZ Center's Spanish-Basque and English-Basque MT models}\footnote{No Basque-French MT system was available from the HiTZ Center at the time of this study.}. To translate the sentences, we loaded each
model and its corresponding tokenizer from Hugging Face using the Transformers library
and generated outputs with the default settings. 

\paragraph{Proprietary translation services.} As a reference for deployed MT systems, we assessed various proprietary translation services. Such systems provide limited transparency with respect to their internal architectures. We included \textbf{Google Translate}, the widely used commercial MT service developed by Google that covers over 100 languages (accessed through the Cloud Translation API); \textbf{Elia}\footnote{\url{https://elia.eus/itzultzailea}} developed by Elhuyar Foundation and supporting Basque, Spanish, French, English, Catalan and Galician; \textbf{Batua}\footnote{\url{https://www.batua.eus/}}, created by Vicomtech and covering Basque, Spanish, French and English; and \textbf{Itzuli}\footnote{\url{https://www.euskadi.eus/itzuli/}}, provided by the Basque Government and supporting the same languages as Batua. Translations with Elia, Batua, and Itzuli were obtained manually through their public web interfaces.

\section{Results}


\paragraph{Moderate alignment with labour distributions.} This analysis of the \wino{} experiments assesses the extent to which model outputs reflect the actual occupational gender distribution. When translating from Basque into Spanish, \textbf{GPT~5}, \textbf{NLLB-200-3.3B} and \textbf{Latxa~3.1~70B~Instruct} show the highest correlation (\(r>0.4\), \(p<0.01\)), which indicates a moderate positive relationship between the gendered forms produced by the models and the real-world gender distribution of professions (see Tab.~\ref{tab:wino-pearson}). For Basque–French translations, correlations are generally lower, although the ranking of systems remains broadly similar. This drop may be due to cross-linguistic differences in gender marking or to disparities in the training data available for French. A closer look at the Llama-based models reveals that \textbf{Latxa}'s adaptation to Basque notably improves alignment compared to its \textbf{Llama~3.1} counterparts of the same size. This suggests that targeted language adaptation can enhance a model's sensitivity to gender patterns present in specific linguistic contexts. More broadly, models explicitly trained or fine-tuned for translation tasks (e.g., NLLB-200-3.3B, SalamandraTA~7B~Instruct, GPT~5) tend to achieve higher correlations than purely general-purpose systems, while smaller instruction-tuned LLMs such as Latxa~3.1~8B and Llama~3.1~8B show weaker alignment.

\begin{table}
    \centering
    \begin{tblr}{
        colspec={Xrlrl},
        cells={font=\footnotesize},
        row{1}={font=\footnotesize\bfseries},
        column{2,4}={rightsep=0pt},
        column{3,5}={leftsep=1pt},
        rowsep=1pt
    }
    \toprule
     & \SetCell[c=2]{c} EU-ES && \SetCell[c=2]{c} EU-FR & \\
    \midrule
    Llama 3.1 8B Instruct     & 0.066 &                       & 0.158 &                       \\
    Latxa 3.1 8B Instruct     & 0.270 & \textsuperscript{**}  & 0.311 & \textsuperscript{***} \\
    Llama 3.1 70B Instruct    & 0.215 & \textsuperscript{*}   & 0.254 & \textsuperscript{**}  \\
    Latxa 3.1 70B Instruct    & 0.409 & \textsuperscript{***} & 0.352 & \textsuperscript{***} \\
    DeepSeek-V3.2-Exp         & 0.259 & \textsuperscript{**}  & 0.275 & \textsuperscript{**}  \\
    Claude 4 Sonnet           & 0.191 & \textsuperscript{*}   & 0.166 &                       \\
    GPT 5            & \textbf{0.487} & \textsuperscript{***} & 0.424 & \textsuperscript{***} \\
    \midrule
    SalamandraTA 7B Instruct  & 0.276 & \textsuperscript{**}  & 0.354 & \textsuperscript{***} \\
    \midrule
    MADLAD-400-3B-MT           & 0.385 & \textsuperscript{***} & 0.195 & \textsuperscript{*}   \\
    NLLB-200-3.3B              & 0.440 & \textsuperscript{***} & \textbf{0.430} & \textsuperscript{***} \\
    HiTZ MT eu-es             & 0.369 & \textsuperscript{***} & \SetCell[c=2]{c} -- &         \\
    \midrule
    Google Translate          & 0.390 & \textsuperscript{***} & 0.359 & \textsuperscript{***} \\
    Batua                     & 0.320 & \textsuperscript{***} & 0.280 & \textsuperscript{**}  \\
    Itzuli                    & 0.284 & \textsuperscript{**}  & 0.258 & \textsuperscript{**}  \\
    Elia                      & 0.246 & \textsuperscript{**}  & 0.243 & \textsuperscript{**}  \\
    \bottomrule
\end{tblr}

\caption{\label{tab:wino-pearson} Pearson correlation coefficients (\( r \)) between Lanbide's labour statistics and model translations from \wino{} to Spanish and French. The highest correlation per language pair is highlighted in bold. \textsuperscript{***}$p<0.01$, \textsuperscript{**}$p<0.05$, \textsuperscript{*}$p<0.1$}
\end{table}

\begin{table}
\centering
\small
\begin{tabular}{lcrr}
\toprule
\multirow{2}{*}{\thead{~\\\textbf{Occupation}}} & \multirow{2}{*}{\thead{\textbf{Real female}\\\textbf{proportion (\%)}}} & \multicolumn{2}{c}{\textbf{GRAPE-M}} \\
\cmidrule(lr){3-4}
 &  & \multicolumn{1}{c}{\textbf{ES}} & \multicolumn{1}{c}{\textbf{FR}} \\
\midrule
housekeeper & 96.5 & 23.4 & 24.6 \\
tailor & 92.7 & 12.4 & 12.5 \\
receptionist & 89.3 & 7.7 & 7.9 \\
cashier & 86.8 & 6.9 & 6.6 \\
hairdresser & 86.2 & 5.9 & 5.9 \\
\bottomrule
\end{tabular}
\caption{
Top five most feminized occupations in the real-world data showing the highest masculine bias scores (GRAPE-M) across all translation models.
}
\label{tab:grape-masculine}
\end{table}

\paragraph{Masculine overrepresentation.}
To complement the correlation analysis, we next examined the direction and magnitude of gender bias using the GRAPE metric. Results from the \wino{} experiment show that all models systematically favour masculine realizations of occupations, both in Spanish and French. The occupation most consistently overrepresented as masculine is \textit{housekeeper}, despite being highly feminized in the Basque labour market (96.5\% female). In Spanish, \textbf{Claude~4~Sonnet}, \textbf{Elia}, and \textbf{DeepSeek-V3.2-Exp} deviate the most from the expected gender distribution for this profession, whereas \textbf{Latxa~3.1~70B}, \textbf{NLLB-200-3.3B}, and \textbf{GPT~5} produce more faithful translations. Other highly feminized professions---such as \textit{tailor}, \textit{receptionist}, \textit{cashier}, \textit{hairdresser}, and \textit{salesperson}---also exhibit strong masculine preferences, with similar GRAPE-M values in both target languages (see Tab.~\ref{tab:grape-masculine}). Instances of feminine overrepresentation are rare. Only ten cases display GRAPE-F values above 1.0, and most correspond to translation errors or lexical ambiguities. For example, in Spanish, \textbf{MADLAD-400-3B-MT} attains the highest GRAPE-F score (3.09) for \textit{ertzain} (\textit{Basque police officer}), often mistranslated as \textit{la Ertzaintza}, referring to the institution rather than an individual. The only occupation that consistently appears predominantly in the feminine form across several models is \textit{nurse}, with GRAPE-F values of 0.18 in Spanish and 0.16 in French. However, these scores remain close to zero, feminine overrepresentation being very weak compared to the masculine one observed earlier. 

\paragraph{Limited gender effects on translation quality.}
Using \flores, we evaluated whether translation quality differs across masculine and feminine sentence variants. We computed the average difference in translation quality between masculine and feminine subsets using the chrF++ and TER metrics and assessed the statistical significance of the observed differences through paired approximate randomization. At first glance, results in Tab.~\ref{tab:flores-results} suggest that, in general, models translate better when the referent is masculine in the Spanish-Basque direction. Nevertheless, most differences are small and not significant. The only exception is \textbf{Batua}, which exhibits significant differences in both metrics, consistently achieving better scores on masculine sentences. On the contrary, in the case of English as the source language, \textbf{NLLB-200-3.3B} shows a significant difference in the opposite direction in terms of chrF++, suggesting slightly better performance on feminine sentences.

\begin{table}[t]
\centering
\small
\resizebox{\linewidth}{!}{%
\begin{tabular}{lrrrr}
\toprule
 & \multicolumn{2}{c}{\textbf{ES-EU}} & \multicolumn{2}{c}{\textbf{EN-EU}} \\
\cmidrule(lr){2-3} \cmidrule(lr){4-5}
 & \multicolumn{1}{c}{\textbf{chrF++}} & \multicolumn{1}{c}{\textbf{TER}} & \multicolumn{1}{c}{\textbf{chrF++}} & \multicolumn{1}{c}{\textbf{TER}} \\
\midrule
Llama 3.1 8B Instruct  &   \cellcolor{green!15}0.22 & \cellcolor{orange!25}-21.07 & 0.00 & \cellcolor{orange!25}-60.71 \\
Latxa 3.1 8B Instruct  &   \cellcolor{green!15}0.14 & \cellcolor{orange!25}-0.50 & \cellcolor{orange!25}-0.11 & \cellcolor{green!15}0.60 \\
Llama 3.1 70B Instruct &   \cellcolor{green!15}0.20 &  \cellcolor{green!15}3.01 & \cellcolor{orange!25}-0.42 &  \cellcolor{orange!25}-0.44 \\
Latxa 3.1 70B Instruct &   \cellcolor{green!15}0.14 &  \cellcolor{green!15}0.11 & \cellcolor{orange!25}-0.22 & \cellcolor{orange!25}-0.44 \\
DeepSeek-V3.2-Exp      & \cellcolor{orange!25}-0.24 & \cellcolor{orange!25}-0.29 & \cellcolor{orange!25}-0.59 &  \cellcolor{green!15}1.22 \\
Claude Sonnet 4        & \cellcolor{orange!25}-0.12 &  \cellcolor{green!15}0.16 & \cellcolor{orange!25}\textbf{-1.08$^{*}$} & \cellcolor{orange!25}-0.53 \\
GPT 5                  & \cellcolor{orange!25}-0.36 &  \cellcolor{green!15}0.50 & \cellcolor{orange!25}-0.25 &  \cellcolor{green!15}0.92 \\
\midrule
SalamandraTA 7B Instruct  &  \cellcolor{green!15}0.20 & \cellcolor{orange!25}-0.33 & \cellcolor{orange!25}-0.03 & \cellcolor{orange!25}-0.36 \\
\midrule
MADLAD-400-3B-MT  &  \cellcolor{green!15}0.19 &  \cellcolor{green!15}0.35 & \cellcolor{orange!25}-0.13 & \cellcolor{orange!25}-1.13 \\
NLLB-200-3.3B    & \cellcolor{orange!25}-0.49 & \cellcolor{orange!25}-0.05 & \cellcolor{orange!25}\textbf{-1.50$^{*}$} & \cellcolor{orange!25}-0.87 \\
HiTZ MT   &  \cellcolor{green!15}0.24 &  \cellcolor{green!15}0.22 & \cellcolor{orange!25}-0.14 &  \cellcolor{green!15}1.27 \\
\midrule
Google Translate     &  \cellcolor{green!15}0.26 &  \cellcolor{green!15}\textbf{0.85$^{*}$} &  \cellcolor{green!15}0.01 &  \cellcolor{green!15}0.77 \\
Batua          &  \cellcolor{green!15}\textbf{0.43$^{*}$} &  \cellcolor{green!15}\textbf{0.97$^{*}$} & \cellcolor{orange!25}-1.00 & \cellcolor{orange!25}-0.82 \\
Itzuli         &  \cellcolor{green!15}0.22 &  \cellcolor{green!15}0.63 & \cellcolor{orange!25}-0.28 &  \cellcolor{green!15}0.28 \\
Elia           &  \cellcolor{green!15}0.17 &  \cellcolor{green!15}0.48 & \cellcolor{orange!25}-0.29 &  \cellcolor{green!15}0.45 \\
\bottomrule
\end{tabular}
}
\caption{Average difference ($\Delta$) between masculine and feminine contrastive sets from \flores. Positive values (\colorbox{green!15}{green}) indicate higher performance on the masculine set, and negative values (\colorbox{orange!25}{orange}) indicate higher performance on the feminine set. Statistically significant results are marked in boldface. \textsuperscript{*}$p<0.05$.}
\label{tab:flores-results}
\end{table}

\paragraph{Impact of annotated linguistic phenomena.}
To carry out a more detailed analysis, we used the annotations of the dataset to explore whether gender-related patterns observed persist under different conditions, such as the presence of multiple gendered entities, gendered proper names or the use of unmarked masculine forms. The values reported in Tab.~\ref{tab:flores-labels} correspond to the difference between masculine–feminine performance when each factor is present compared to when it is not.

In our analyses, most gender-related differences were not statistically significant; however, several systems show significant or systematic tendencies under specific conditions. When sentences include multiple gendered entities, \textbf{HiTZ MT} tends to yield higher scores for masculine contexts in either language pair (e.g., \textit{Es poco probable que \ul{el conductor}}\textsubscript{ \male} \textit{del vehículo que embistió \ul{al fotógrafo}}\textsubscript{ \male} \textit{sea procesado penalmente}, or \textit{\ul{Liggins}}\textsubscript{ \male} \textit{followed in \ul{his}}\textsubscript{ \male} \textit{\ul{father}}\textsubscript{ \male}\textit{'s footsteps and entered a career in medicine}). \textbf{Itzuli} also achieves significantly better (lower TER) performance in masculine multi-entity settings when translating from Spanish. Regarding proper names, \textbf{Elia} and \textbf{NLLB-200-3.3B} perform significantly better with feminine referents when translating from Spanish, as in \textit{Después de eso, \ul{Kavita Paudwal}}\textsubscript{ \female} \textit{pasó al frente para cantar los bhajans}. Other systems show no clear advantage for either gender. In contrast, when translating from English, four models---\textbf{Latxa-70B}, \textbf{DeepSeek-V3.2-Exp}, \textbf{Batua} and \textbf{Itzuli}--- show significant advantages for masculine names. Among the examined variables, the use of unmarked masculine (only applicable in Spanish) had the most consistent impact. Systems like \textbf{NLLB-200-3.3B}, \textbf{Itzuli} and \textbf{Elia} achieved better results for sentences containing unmarked masculine forms compared to their feminine counterparts  (e.g., \textit{\ul{Los parisinos}}\textsubscript{ \male} \textit{tienen reputación de egocéntricos, descorteses y engreídos}).

\begin{table*}[t]
\centering
\small
\resizebox{\textwidth}{!}{
\begin{tabular}{@{}lrrrrrrrrrr@{}}
\toprule
 &
  \multicolumn{6}{c}{\textbf{ES-EU}} &
  \multicolumn{4}{c}{\textbf{EN-EU}} \\ 
  \cmidrule(lr){2-7} \cmidrule(lr){8-11}
&
  \multicolumn{2}{c}{\textbf{ME}} &
  \multicolumn{2}{c}{\textbf{PN}} &
  \multicolumn{2}{c}{\textbf{UM}} &
  \multicolumn{2}{c}{\textbf{ME}} &
  \multicolumn{2}{c}{\textbf{PN}} \\ 
  \cmidrule(lr){2-3} \cmidrule(lr){4-5} \cmidrule(lr){6-7} \cmidrule(lr){8-9} \cmidrule(lr){10-11}
 &
  \multicolumn{1}{c}{\textbf{chrF++}} &
  \multicolumn{1}{c}{\textbf{TER}} &
  \multicolumn{1}{c}{\textbf{chrF++}} &
  \multicolumn{1}{c}{\textbf{TER}} &
  \multicolumn{1}{c}{\textbf{chrF++}} &
  \multicolumn{1}{c}{\textbf{TER}} &
  \multicolumn{1}{c}{\textbf{chrF++}} &
  \multicolumn{1}{c}{\textbf{TER}} &
  \multicolumn{1}{c}{\textbf{chrF++}} &
  \multicolumn{1}{c}{\textbf{TER}} \\
\midrule
Llama 3.1 8B Instruct &
  \cellcolor{green!15}0.44 & \cellcolor{orange!25}-130.60 &
  \cellcolor{orange!25}-0.28 & \cellcolor{orange!25}-33.64 &
  \cellcolor{green!15}1.00 & \cellcolor{green!15}34.65 &
  \cellcolor{green!15}1.02 & \cellcolor{orange!25}-65.17 & 
  \cellcolor{green!15}2.06 & \cellcolor{green!15}0.61 \\ 
Latxa 3.1 8B Instruct &
  \cellcolor{orange!25}-0.74 & \cellcolor{green!15}0.17 &
  \cellcolor{orange!25}-0.33 & \cellcolor{orange!25}-1.74 &
  \cellcolor{green!15}0.07 & \cellcolor{green!15}1.29 &
  \cellcolor{green!15}0.65 & \cellcolor{orange!25}-0.75 & 
  \cellcolor{green!15}0.43 & \cellcolor{green!15}0.75 \\  
Llama 3.1 70B Instruct &
  \cellcolor{green!15}0.08 & \cellcolor{orange!25}-28.39 &
  \cellcolor{orange!25}-0.26 & \cellcolor{orange!25}-31.92 &
  \cellcolor{green!15}0.07 & \cellcolor{orange!25}-6.83 &
  \cellcolor{green!15}\textbf{1.57$^{*}$} & \cellcolor{green!15}2.52 &
  \cellcolor{green!15}0.92 & \cellcolor{green!15}2.84 \\ 
Latxa 3.1 70B Instruct &
  \cellcolor{orange!25}-0.28 & \cellcolor{green!15}0.31 &
  \cellcolor{green!15}0.08 & \cellcolor{green!15}0.43 &
  \cellcolor{orange!25}-0.12 & \cellcolor{orange!25}-0.24 &
  \cellcolor{green!15}1.79 & \cellcolor{green!15}2.22 & 
  \cellcolor{green!15}\textbf{1.87$^{*}$} & \cellcolor{green!15}2.73 \\
DeepSeek-V3.2-Exp &
  \cellcolor{green!15}0.80 & \cellcolor{green!15}2.03 &
  \cellcolor{green!15}0.21 & \cellcolor{green!15}1.52 &
  \cellcolor{orange!25}-0.26 & \cellcolor{orange!25}-0.81 &
  \cellcolor{white}0.00 & \cellcolor{green!15}1.32 & 
  \cellcolor{green!15}1.08 & \cellcolor{green!15}\textbf{4.23$^{**}$} \\
Claude Sonnet 4 &
  \cellcolor{green!15}0.38 & \cellcolor{white}0.00 &
  \cellcolor{orange!25}-0.46 & \cellcolor{orange!25}-0.94 &
  \cellcolor{green!15}0.12 & \cellcolor{green!15}0.27 &
  \cellcolor{green!15}1.34 & \cellcolor{green!15}0.91 & 
  \cellcolor{orange!25}-1.14 & \cellcolor{orange!25}-1.27 \\
GPT 5 &
  \cellcolor{green!15}0.53 & \cellcolor{green!15}0.19 &
  \cellcolor{orange!25}-0.36 & \cellcolor{orange!25}-0.17 &
  \cellcolor{green!15}0.54 & \cellcolor{green!15}1.23 &
  \cellcolor{green!15}1.55 & \cellcolor{green!15}1.62 &
  \cellcolor{orange!25}-0.44 & \cellcolor{orange!25}-0.30 \\
\midrule
SalamandraTA 7B Instruct &
  \cellcolor{orange!25}-0.10 & \cellcolor{green!15}1.45 &
  \cellcolor{orange!25}-0.55 & \cellcolor{orange!25}-0.68 &
  \cellcolor{green!15}0.45 & \cellcolor{orange!25}-0.55 &
  \cellcolor{orange!25}-0.14 & \cellcolor{green!15}1.64 & 
  \cellcolor{orange!25}-1.68 & \cellcolor{orange!25}-0.80 \\ 
\midrule
MADLAD-400-3B-MT &
  \cellcolor{green!15}0.42 & \cellcolor{green!15}1.17 &
  \cellcolor{orange!25}-0.62 & \cellcolor{orange!25}-2.23 &
  \cellcolor{green!15}0.24 & \cellcolor{green!15}0.26 &
  \cellcolor{green!15}0.61 & \cellcolor{green!15}2.09 & 
  \cellcolor{orange!25}-1.12 & \cellcolor{orange!25}-2.42 \\ 
NLLB-200-3.3B &
  \cellcolor{orange!25}-0.07 & \cellcolor{green!15}0.36 &
  \cellcolor{orange!25}\textbf{-1.44$^{*}$} & \cellcolor{orange!25}-1.09 &
  \cellcolor{green!15}\textbf{1.88$^{**}$} & \cellcolor{green!15}\textbf{2.05$^{**}$} &
  \cellcolor{green!15}0.59 & \cellcolor{white}0.00 &  
  \cellcolor{green!15}-0.19 & \cellcolor{green!15}2.16 \\ 
HiTZ MT &
  \cellcolor{green!15}\textbf{0.84$^{**}$} & \cellcolor{green!15}0.58 &
  \cellcolor{orange!25}-0.34 & \cellcolor{orange!25}-1.12 &
  \cellcolor{green!15}0.56 & \cellcolor{green!15}1.01 &
  \cellcolor{green!15}\textbf{2.27$^{**}$} & \cellcolor{green!15}\textbf{4.53$^{**}$} &  
  \cellcolor{green!15}0.80 & \cellcolor{green!15}2.00 \\
\midrule
Google Translate &
  \cellcolor{orange!25}-0.31 & \cellcolor{green!15}0.93 &
  \cellcolor{green!15}0.36 & \cellcolor{green!15}0.66 &
  \cellcolor{green!15}0.30 & \cellcolor{green!15}0.47 &
  \cellcolor{green!15}1.59 & \cellcolor{green!15}1.58 & 
  \cellcolor{green!15}0.85 & \cellcolor{green!15}2.22 \\
Batua &
  \cellcolor{white}0.00 & \cellcolor{orange!25}-0.07 &
  \cellcolor{green!15}0.10 & \cellcolor{orange!25}-0.52 &
  \cellcolor{orange!25}-0.41 & \cellcolor{orange!25}-0.08 &
  \cellcolor{green!15}1.27 & \cellcolor{green!15}2.99 & 
  \cellcolor{green!15}0.58 & \cellcolor{green!15}\textbf{3.30$^{*}$} \\ 
Itzuli &
  \cellcolor{green!15}0.50 & \cellcolor{green!15}\textbf{1.59$^{*}$} &
  \cellcolor{orange!25}-0.42 & \cellcolor{green!15}0.73 &
  \cellcolor{green!15}\textbf{1.05$^{**}$} & \cellcolor{orange!25}0.33 &
  \cellcolor{green!15}0.33 & \cellcolor{green!15}2.00 & 
  \cellcolor{green!15}0.15 & \cellcolor{green!15}\textbf{2.38$^{*}$} \\ 
Elia &
  \cellcolor{orange!25}-0.21 & \cellcolor{orange!25}-1.18 &
  \cellcolor{orange!25}\textbf{-1.00$^{*}$} & \cellcolor{orange!25}\textbf{-2.43$^{**}$} &
  \cellcolor{green!15}\textbf{1.19$^{**}$} & \cellcolor{green!15}\textbf{1.89$^{**}$} &
  \cellcolor{green!15}1.54 & \cellcolor{green!15}2.53 & 
  \cellcolor{green!15}0.78 & \cellcolor{green!15}1.05 \\ 
\bottomrule
\end{tabular}
}
\caption{Interaction effect ($\Delta$) between gender and linguistic factors (ME = Multi-entity, PN = Proper names, UM = Unmarked masculine) across both Spanish-Basque (ES-EU) and English-Basque (EN-EU) translation directions. \colorbox{green!15}{Green} indicates higher performance on masculine sentences, \colorbox{orange!25}{orange} indicates higher performance on feminine ones. Statistically significant results are marked in boldface. \textsuperscript{**}$p<0.05$, \textsuperscript{*}$p<0.1$}
\label{tab:flores-labels}
\end{table*}

\section{Discussion}

The results indicate that current MT systems and LLMs continue to reproduce gender asymmetries. In translation from Basque into gendered target languages, all evaluated systems show a systematic preference for masculine realizations of gender-neutral occupational terms. This behaviour is consistent with earlier reports of related work on other language pairs~\citep{gete-etchegoyhen-2024-context,mastromichalakis2025assumedidentitiesquantifyinggender,vanmassenhove-etal-2018-getting,Schiebinger2014ScientificRM,monti2020gender} and reflects the grammatical conventions of target languages where masculine forms function as the unmarked default, such as Spanish and French.

Although all models display a systematic preference for masculine realizations, the degree of bias varies across occupations. Professions that are predominantly feminine in real data (e.g., \textit{housekeeper}, \textit{tailor}, \textit{receptionist}) are still often translated in the masculine---with a few exceptions, notably \textit{nurse}---, yet the moderate correlations with Basque labour statistics indicate that models do capture some aspects of real-world gender distributions. In other words, while translation outputs reflect existing occupational asymmetries to a limited extent, they systematically exaggerate the masculine default due to its higher frequency and unmarked grammatical status in the training data.


The analysis of the translation \emph{into} Basque using \flores{} revealed a weaker and less systematic gender effect. For Spanish sources, most models produced slightly but rarely significant higher scores for masculine variants, particularly in sentences containing the generic masculine (e.g. \textit{los investigadores}, `the researchers') compared to their marked feminine counterparts (\textit{las investigadoras}, `the \emph{female} researchers'). These results mirror the bias observed above, and suggest that the unmarked status and higher frequency of masculine forms in Spanish data may influence translation quality. In contrast, translations from English, a weakly gendered language, showed no consistent direction of bias: a few models achieved marginally better results for feminine sentences, while other performed slightly better with masculine proper names or multi-entity contexts.

\section{Conclusions}

This work introduces two high-quality resources for evaluating gender bias in translation involving Basque, a low-resource and genderless language. \wino{} adapts the WinoMT benchmark to Basque and provides a manually post-edited and culturally localised dataset that enables systematic assessment of how gender-neutral mentions of occupations are rendered in gendered target languages. Further, each occupation mention in \wino{} has been manually aligned with official Basque labour statistics, allowing researches to compare the gender distributions that emerge in model-generated translations against real-world employment data. This design makes it possible to quantify how translation systems reflect or diverge from actual societal gender patterns, an evaluation dimension rarely supported in previous resources.

\flores{}, in turn, extends the FLORES+ benchmark with manually annotated gender-controlled contrastive pairs in Spanish and English. Each sentence has been validated for semantic equivalence and enriched with linguistic annotations (i.e., multiple entities, proper names, and unmarked masculine), creating a valuable testbed for analysing how grammatical gender in the source text affects translation quality into Basque.


Our experiments confirmed that current MT and LLM systems still default to masculine forms, even when the occupations are clearly associated with women. Regarding translation quality, results were less consistent: translations from Spanish, a strongly-gendered language, tended to be slightly better for masculine referent in some models, while no clear pattern emerged for English, a weakly-gendered language. Future work will leverage these resources to compare bias patterns across sociocultural contexts and to inform bias-aware training and evaluation practices involving the Basque language.


\section{Acknowledgements}

This work was supported by the HiTZ Chair of Artificial Intelligence and Language Technology
(TSI100923-2023-1), funded by MTDFP, \textit{Secretaría de Estado de Digitalización e Inteligencia Artificial}. Additional support was provided by the Research Project PID2024-157855OB-C32 (MOLVI), funded by MICIU/AEI/10.13039/501100011033 and the European Regional Development
Fund (ERDF), EU. It was also funded by the Basque Government (IKER-GAITU project) and the \textit{Ministerio para la Transformación Digital y de la Función Pública} - Funded by EU – NextGenerationEU within the framework of the project \textit{Desarrollo de Modelos ALIA}.

\section{Ethics Statement and Limitations}
Our work has several limitations that should be taken into account. First of all, the LLMs we evaluated show considerable variability, that is to say, small modifications in prompt design can lead to notable differences in the generated translations. Additionally, these models are typically used with relatively high values of temperature, which further increases randomness in their outputs. Fixing temperature at 0 allows to reproduce the experiments, but fails to reflect the variability present in real-world applications.

Regarding the newly created datasets, on the one hand, \wino{} focuses on a specific type of representational harm---misrepresentation---and covers a narrow domain, occupational representations, which captures only one aspect of gender bias in language technologies. Furthermore, this dataset is template-based, which is useful for controlled experimentation, but it does not represent the real use of language. \flores, on the other hand, is based on real texts, but still presents some limitations. The size of both English and Spanish gendered sets is relatively small, and this may limit the reliability of the results. Apart from the size, this dataset also includes proper names to infer gender, which must be used carefully, given that relying on personal names without critical consideration can introduce or reinforce unintended biases. Translation quality was assessed through automatic metrics, which may not reflect the nuances of gender-related differences.

Another important limitation is that our approach treats gender as a binary variable (masculine/feminine). Although this choice reflects the grammatical systems of the analysed languages, it disregards non-binary and gender-neutral identities and provides an oversimplified view of gender in language. 

Lastly, it should be noted that bias is a complex phenomenon that can manifest in many different ways depending on the task, domain or social context in which the models are used. Therefore, we cannot rely on a single NLP task, bias domain, metric or dataset to determine whether a model is biased. A high or low bias score in our datasets only should not be interpreted as evidence of generalized bias or the absence of it, but as an indication of how the model behaves in specific tasks and experimental conditions.

\section{Bibliographical References}\label{sec:reference}

\bibliographystyle{lrec2026-natbib}
\bibliography{bibliografia}


\end{document}